\documentclass{article}

     \usepackage[final]{neurips_2020}

     \usepackage{neurips_2020}

\usepackage[utf8]{inputenc} 
\usepackage[T1]{fontenc}    
\usepackage{hyperref}       
\usepackage{url}            
\usepackage{booktabs}       
\usepackage{amsfonts}       
\usepackage{nicefrac}       
\usepackage{microtype}      
\usepackage{graphicx}
\usepackage{amsmath}

\newcommand\context{\boldsymbol{e}(w_{<t})}
\newcommand\att{\boldsymbol{att}}
\newcommand\act{\boldsymbol{act}}
\newcommand\ctx{\boldsymbol{ctx}}
\newcommand\compemb{\boldsymbol{ce}^{i}}

\title{Neural Composition: Learning to Generate from Multiple Models}

\author{%
Denis Filimonov\\
\texttt{denf@amazon.com}
\And Ravi Teja Gadde\\
\texttt{gadderav@amazon.com}
\And Ariya Rastrow\\
\texttt{arastrow@amazon.com}
}

\begin{document}

\maketitle

 \setcitestyle{numbers,square}
 
\begin{abstract}
Decomposing models into multiple components is critically important in many applications such as
            language modeling (LM)   as it enables adapting individual components separately
            and biasing of some components to the user's personal preferences.
Conventionally, contextual and personalized adaptation for language models, are achieved through class-based factorization,
	which requires class-annotated data, or through biasing to individual phrases which is limited in scale. 
In this paper, we propose a system that combines \emph{model-defined} components, by learning when to activate the generation process from each individual component, 
	and how to combine probability distributions from each component, directly from unlabeled text data.
\end{abstract}

\section{Introduction}
\label{section:intro}
Language models are a key component of applications that require generation of coherent natural language text, 
	including machine translation, speech recognition, abstractive text summarization, and many others.
For a long time n-gram models \citep{ueberla1993ngram} dominated the field due to their simplicity, efficiency and scalability.
However, recently neural models gained popularity, notably from simple recurrent networks \citep{mikolov2010rnnlm} to 
	very powerful models including \citep{gpt2, yang2019xlnet}.
These models often include billions of parameters and they have been shown to do very well at generalizing from vast amounts of data.
However, how to \emph{adapt} these models to different users (e.g., personalized contact list in a messaging application), 
	or how to update these models efficiently (for example, when a new movie title is released, which may be important for a ticket booking application)
	does still remain a challenge. 
When the number of users is large, or updates are frequent, adapting a large monolithic model becomes impractical 
	and this necessitates the use of composite models in which some components may be updated separately.

For these reasons, class-based models are still widely used in different applications, particularly in automatic speech recognition (ASR) where integrating external knowledge sources and personalized entities in the language model are crucial in achieving accurate transcription:
	\citep{contactnames2015, dynclass2016vasserman, mcgraw2016personalized, chen2018endtoend}.
Class-based models, however, require annotations in order to learn where these components/classes are used which limits their applicability.
Instead of using classes, where content of a class is assumed to be similar in some way, e.g., entities of the same type, 
	\citep{hall2015ngram, scheiner2016voicesearch, he2019e2e} boost scores of individual phrases and n-grams to bias ASR search.
Note that this type of biasing can be applied to both WFST-based\footnote{Weighted finite-state transducers (WFSTs) are widely used in speech recognition to represent language models \citep{mohri2002wfst}.} and neural models. 


\citep{pundak2018deep} learn a fixed-size representation for every biasing phrase separately.
The ASR decoder then uses attention mechanism to interpolate these representations and the result is added to the decoder's input.
As the decoder needs to attend to each individual phrase at every step, scaling this approach to a large number of biasing phrases and entities poses an engineering challenge. 
\citep{khandelwal2020generalization} propose nearest-neighbor LM which can use external data to bias its predictions,
	however, significant limits application of this type of model, especially in ASR domain.
\citep{levit2015personalization} is similar to our work in that they aim to solve a similar problem.
They use expectation-maximization method to learn a class-based (or more generally, \emph{word-phrase-entity}) model without a requirement for annotated data.
However, their method only applies to n-gram models while we do not make assumptions about internal structure of component models.

In this paper, we take an approach reminiscent of a class-based model in that we use components (classes)
	whose elements are expected to be used in similar context.
	
We call them \emph{model-defined} components because they are defined by their respective models (FST- or neural-based).
 Unlike class-based models, however, we do not assign any tags to these components.
 This allows us to do away with one of the main shortcomings of class-based models -- the requirement for annotated (manually or automatically) data.
 The main motivating idea of our method is as follows:
 	given a general generative language model and some components represented as generative LMs, 
	we can learn where these components are \emph{useful}, i.e. where they make better predictions than the general model.
	Additionally, the proposed model learns, directly from data, how to interpolate different components at each token, which class-based approaches are incapable of due to their explicit factorization into sequence of classes and words. 
Note that our approach does not require us to assign any semantic tags to components,
	their meaning is implicit and arises from their content.

It is worth noting that there are many methods for combining multiple \emph{full-sentence} language model in the literature, \cite{iyer1997ood, kalai1999online, broman2005} to name a few.
However, such methods cannot be applied to \emph{entity} models with full-sentence models, and therefore these methods cannot solve the problem we seek to address.

The rest of the paper is organized as follows:
	in Section~\ref{section:model}, we describe the structure of the proposed model,
	the training procedure is detailed in Section~\ref{section:model-training}.
In Section~\ref{section:experiments}, we present experimental results, 
	and in Section~\ref{section:conclusions} we conclude and outline future work.

\section{Compositional Language Model}
\label{section:model}
\label{section:model-structure}
\begin{figure}[htbp]
\begin{center}
\includegraphics[width=0.8\linewidth]{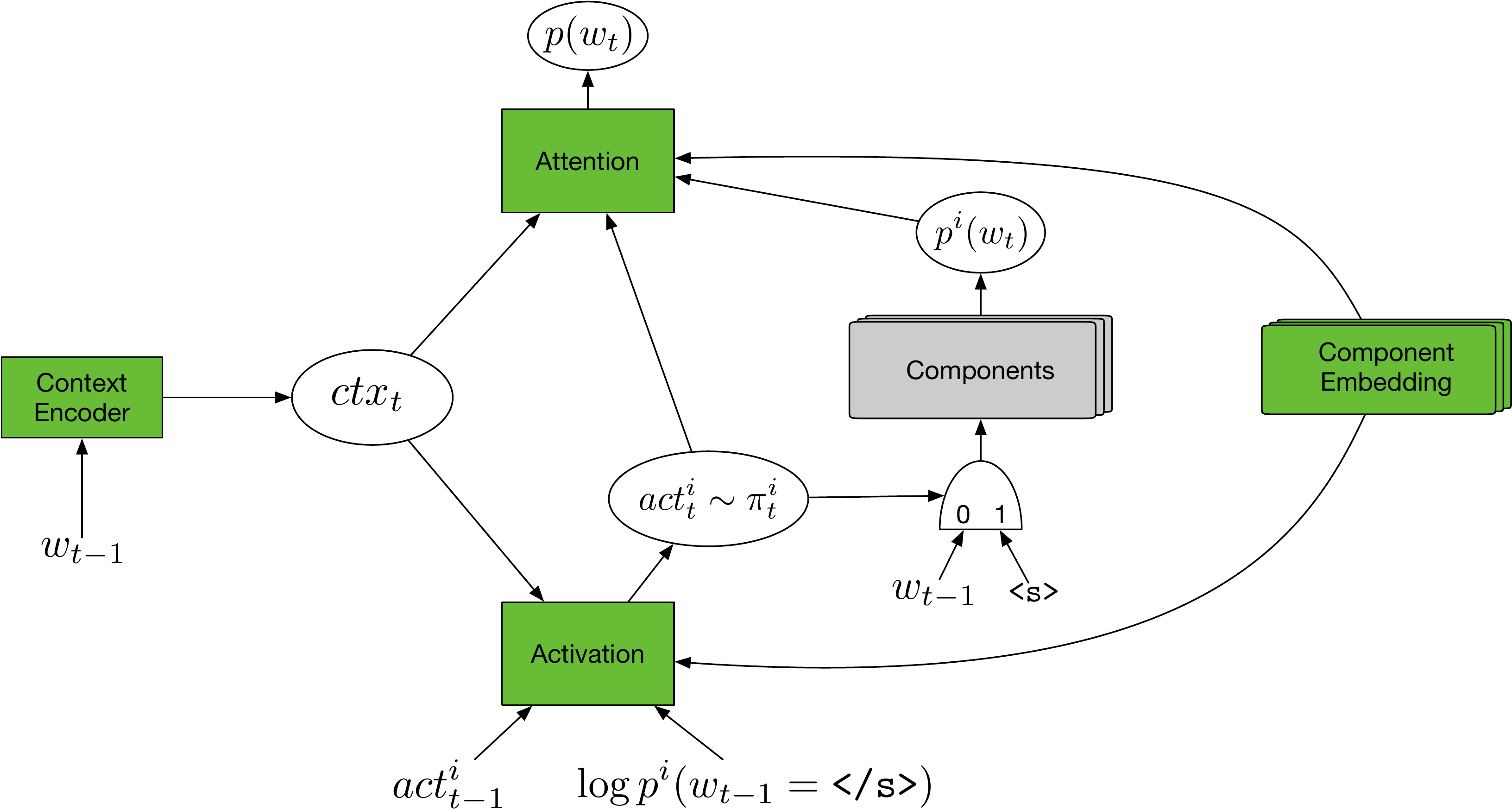}
\caption{Compositional Model Structure. Green color indicates learnable parameters. Component models (in grey) are fixed during the compositional model training.}
\label{fig:model-structure}
\end{center}
\end{figure}

In this section, we present the structure of the proposed compositional model and describe each of its components in detail.
Figure~\ref{fig:model-structure} outlines the structure of the model.
In Sections~\ref{section:components} through \ref{section:attention}, 
	we describe every part of the model in detail, and in Section~\ref{section:model-structure-discussion},
	we explain the rationale: how these parts work together to achieve our goal.

\subsection{Components}
\label{section:components}
Component models are language models, $p^{i}(w_t|w_{<t})$ (we use superscript to indicate component-specific items).
They are implemented as stateful functions ${\cal F}^{i} : {\cal S}^{i}(w_{<t-1}), w_{t-1} \rightarrow {\cal S}^{i}(w_{<t}), p(w_{t})$, with a special start state ${\cal S}^{i}(\emptyset)$.
In this paper, we use LSTM- and FST-based component models.

Each component model is learned independently, the only requirement is shared vocabulary of word (or subword~\citep{tensor2tensor}) tokens.
Some components may model entire sentences while others may model only parts of them, e.g. specific entity types.

\subsection{Default Model}
One component model, component zero, is designated as the \emph{default} model, and it must span the entire sentence.
The default model serves as the baseline model which we aim to adapt and improve by combining with other components.

\subsection{Component Embeddings}
\label{section:component-embeddings}

A component embedding, $\compemb$, is a fixed-size learnable vector associated with each component, and since the components themselves are fixed during the composite model training, this is the only component-specific representation learned by the model. 

\subsection{Context Encoder}
\label{section:context-encoder}
Context encoder maps the variable length word sequence $w_{<t}$ into a fixed-size vector $\ctx_{t}$
	which serves as an input to activation and attention models described below.
In this paper, we use an LSTM network to encode the context $w_{<t}$ with input embedding $\context$.

\begin{equation}
\label{eq:context-encoder}
\ctx_{t} = \mbox{LSTM}(\context)
\end{equation}

\subsection{Activation}
\label{section:activation}
Component models that generate only parts of a sentence, such as entities, are ignorant of the context where those entities can be used, and therefore need an explicit binary signal when to generate their first word, i.e., when to output $p^{i}(w|\verb|<s>|)$.
In the case of an FST-based component model, the activation signal $act^{i}_{t} = 1$ resets the FST state to its start.
When $act^{i}_{t} = 0$ the model generates the next token given its state and $w_{t-1}$.
In the case of an LSTM-based component model, we reset the LSTM state to zero values and replace the previous word $w_{t-1}$ with \verb|<s>| (only for that specific component).

In order to generate the activation signal, we define \emph{activation policy} function $\pi_{t}^{i}$ 
	which can be interpreted as probability of activating component $i$ at time $t$:

\begin{equation}
\label{eq:activation}
\pi_{t}^{i} = \sigma(\mbox{PROJ}(\mbox{LSTM}(\ctx_{t}, \compemb, \boldsymbol{act}_{t-1}^{i}, \log{p^{i}(w_{t-1} = \verb|</s>| | w_{<t-1}, \boldsymbol{act}_{< t}^{i})})))
\end{equation}

\noindent
where the comma indicates concatenation, $\ctx_{t}$ is the context encoding (Eq.~\ref{eq:context-encoder}),
	$\compemb$ is the \emph{component embedding} described in Section~\ref{section:component-embeddings},
	$\act_{t-1}^{i}$ is the binary activation of the component at $t-1$,
	$\log{p^{i}_{t-1}(w = \verb|</s>|)}$ is the log probability of the component generating \verb|</s>| at $t-1$.
Finally, $\mbox{PROJ}$ is a linear projection (with a bias) mapping the LSTM output to scalar input to sigmoid activation function $\sigma$.

For training, we sample binary activations $\act_{t}^{i}$ from $\pi_{t}^{i}$, and for inference we apply a threshold of 0.5.
We describe sampling activations in more detail in Section~\ref{section:lookahead-activation}.

\subsection{Attention}
\label{section:attention}
The role of the attention is to interpolate the outputs of all components:

\begin{equation}
\label{eq:interpolation}
p(w_{t}| w_{<t}, \act_{\le t}^{1 \ldots N}) = \sum_{i=0}^{N} \alpha^{i}_{t} \cdot p^{i}(w_{t} | w_{<t}, \act_{\le t}^{i}),
 ~~\mathrm{ where } ~\alpha^{i}_{t} = \frac{exp(\att^{i}_{t})}{\sum_{i=0}^{N} exp(\att^{i}_{t})}  \mathrm{, and}
\end{equation}

\begin{equation}
\label{eq:attention}
\att^{i}_{t} = \mbox{PROJ}(\mbox{LSTM}(\ctx_{t}, \compemb, \act_{t}^{i}, \log{p^{i}(w_{t} = \verb|</s>| | w_{<t}, \act_{\le t}^{i} )}))
\end{equation}

\noindent
Note that structurally, attention is very similar to activation model in Eq.~\ref{eq:activation}.
The main difference is that
	at a given time $t$, the activation influences input to the component whereas the attention uses its output.
Also note that the activation is computed \emph{independently} for each component (Eq.~\ref{eq:activation}), 
	while the attention coefficients $\alpha^{i}_{t}$ in Eq.~\ref{eq:interpolation} are normalized across all components $i$.
Figure~\ref{fig:example-sentence} shows an example of activation and attention output.

\subsection{Discussion}
\label{section:model-structure-discussion}
Activation and attention networks have similar structure and a related function: 
	they compare the component's embedding against the sentence context captured by the context encoder (Section~\ref{section:context-encoder}).
When a component generated a limited span, within a sentence, it is important to know where the span \emph{starts} and \emph{ends}. Start is learned by the activation model and it is the responsibility of the attention model to learn where a component ends. However, the attention does not have access to the content of the component, therefore, it needs a signal from the component itself.
To that end, we add $\act_{t}^{i}$ and $\log{p^{i}(w_{t} = \verb|</s>| \ldots )}$ to the attention model's inputs.
The activation model gets the same signals but from the previous step $t-1$.
Note that a component predicting \verb|</s>| signals the end of its span but not necessarily the end of sentence which is the same token.
The attention model learns to reduce attention to the component once it starts generating \verb|</s>| with high probability
	until the component is activated again.
This way representing the span on a component is agnostic to its internal structure and works well with both neural and FST-based components.

\begin{figure}[htbp]
\begin{center}
\includegraphics[width=0.8\linewidth]{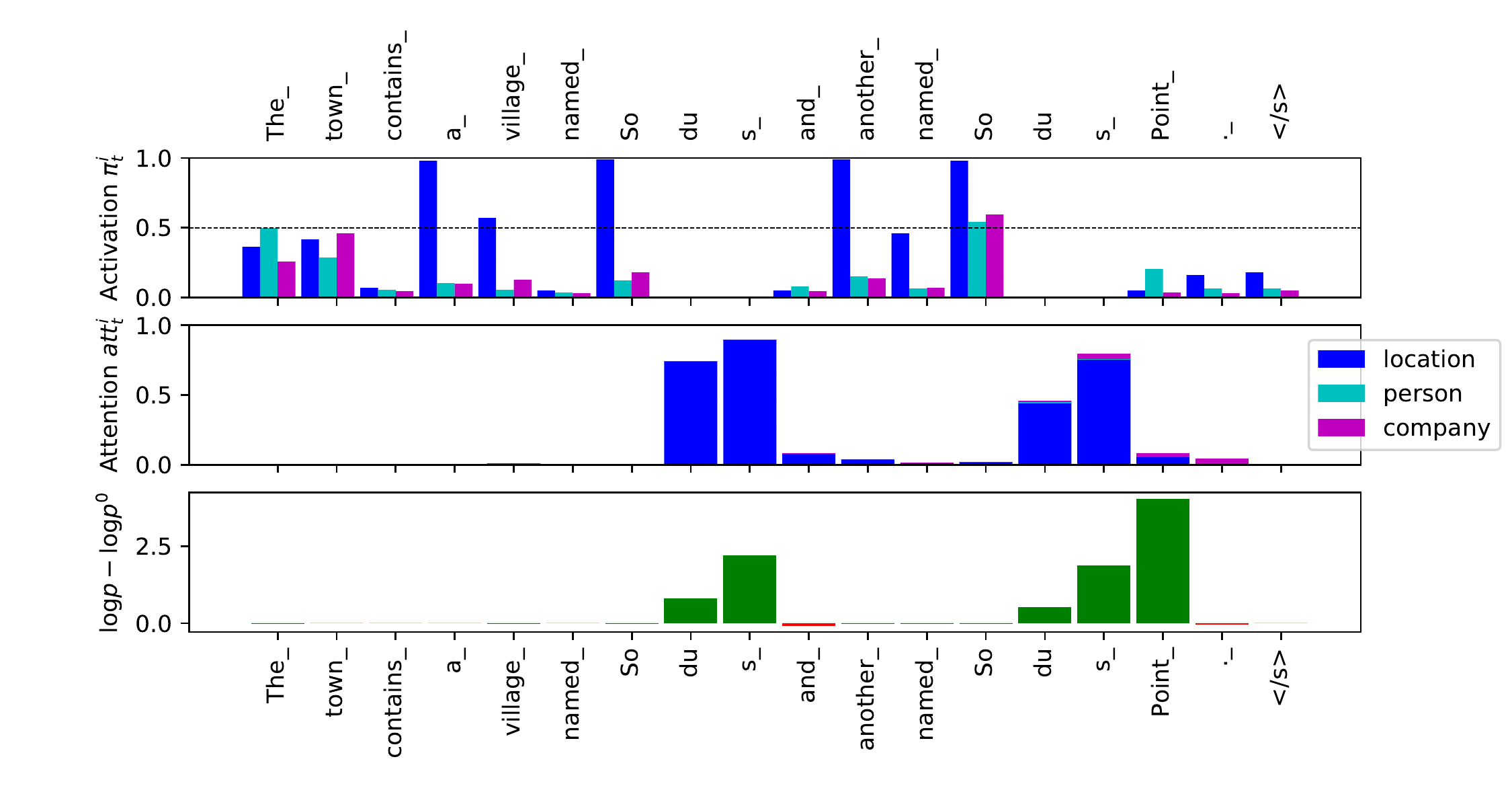}
\caption{This figure illustrates the output of activation and attention models on the sentence \textit{``The town contains a village named Sodus and another named Sodus Point .''} (tokenized into subwords).
}
\label{fig:example-sentence}
\end{center}
\end{figure}

Figure~\ref{fig:example-sentence} illustrates how activation and attention models work together.
In this example, we use 3 components (apart from the default model), representing location, person, and company entity types.
More details about this model are in Section~\ref{section:wikipedia}. Attention is stacked and sums to 1 (including the default mode which is not shown), but activation is independent for each component. We use the threshold of 0.5 to trigger activation.
The bottom graphs shows the difference in token log-likehood between the compositional model and the default model.
Note that the "location" activation spikes at positions where a location is plausible.  However, the attention model moderates the probability mass assigned to this component and also, where the span of location ends.

\section {Model Training}
\label{section:model-training}

If we consider binary activations to be an input, the rest of the model's parameters can be learned by minimizing cross-entropy loss.
Activations can be viewed as actions in a reinforcement learning framework, 
	and we can utilize policy learning to learn activation probability $\pi^{i}_{t}$ (Eq.~\ref{eq:activation}).
In the rest of this section, we provide details on parameter estimation procedure.

\subsection{Loss Functions}
\label{section:loss-functions}

We use a combination of cross entropy (LL) and reinforcement learning (RL) losses\footnote{We omit the conditional part of $p(w_{t})$ for simplicity of notation},

\begin{equation}
\label{eq:LL}
\textbf{LL:}~ \Delta \theta \propto \sum_{t} \frac{\partial \log p(w_{t})}{\partial \theta}
\end{equation}
where $\theta$ refers to all parameters of the model except the activation model. 
The RL loss, on the other hand, aims to maximize expected reward under the activation policy function $E_{\pi^{i}_{t}}[p(w_{t})]$,
	and the parameter update is as follows (REINFORCE algorithm) \citep{williams92reinforce}:

\begin{equation}
\label{eq:reinforce}
\textbf{RL:}~ \Delta \zeta \propto - \sum_{t} G_{t} \frac{\partial \log \pi^i_t }{\partial \zeta}
\end{equation}

\noindent 
where $\pi^{i}_{t}$ is the activation policy (Eq.~\ref{eq:activation}) with parameters $\zeta$.
Here, $G_{t} = \sum_{\tau=t}^{T} R_{\tau}$ is the reward $R_{\tau}$ from time $t$ to the end of the sentence.
We use the following reward function:

\begin{equation}
\label{eq:reward}
R_{\tau} = \log p(w_{\tau}) - \log p^{0}(w_{\tau})
\end{equation}

where $p(w_{\tau})$ and $p^{0}(w_{\tau})$ are the probability of compositional and the default models, respectively. 
This reward is equivalent to minimizing expected cross-entropy of the compositional model but subtracting the default model reduces variance of the update.

REINFORCE update in Eq.~\ref{eq:reinforce} relies on activation samples drawn from $\pi^{i}_{t}$.
We deviate from this, and draw samples from an interpolated policy $\hat{\pi}^{i}_{t}$ instead:

\begin{equation}
\label{eq:interpolated-policy}
\hat{\pi}^{i}_{t} = \lambda b^{i}_{t} + (1 - \lambda) \pi^{i}_{t}
\end{equation}

\noindent
where $b^{i}_{t}$ is a non-parametric behavior activation policy acting as a teacher and $\lambda$ is the interpolation weight 
	which starts from 1 (behavior policy only) and decays by a schedule. In Section~\ref{section:lookahead-activation}, we describe the rationale for using the teacher activation policy and its implementation details. Note that sampling activations from a different policy results in a biased expected cross entropy estimator.
The bias, however, decreases as the training progresses, so we do not correct it.\footnote{Unbiased estimator using importance sampling \citep{hurtado1998monte} results in a high update variance which prevents the model from converging.}

\subsection{Lookahead Teacher Activation}
\label{section:lookahead-activation}

There is a subtle but important nuance about learning attention and activation:
	if the binary activation sequence for a component is too random, the component's output will not be useful,
	thus minimizing cross-entropy loss will cause the attention model to ignore that component.
Conversely, when the attention to a component is either too random or too small, 
	the reward function (Eq.~\ref{eq:reward}) becomes indifferent to changes in activation, which prevents the activation model from learning.
This is a chicken-and-egg problem: we need a good activation policy to learn attention and we need a good reward function (attention) to learn activation.
We resolve this problem by introducing a lookahead teacher activation policy.

Note that during training we know the entire sentence ahead of time, so if we activate a component at a certain time,
	we can compare the likelihood of generating subsequent words under the component and the default model.
Specifically,

\[
b^{i}_{t} \propto \frac{p^{i}(w_{t} | act^{i}_{t} = 1)}{p^{0}(w_{t})}
\]

Note that this is a non-parametric policy as all components are fixed when we train the composite model.
It is possible to extend the lookahead to multiple words in the future, but in our experiments one word was sufficient.
However, in our subword models, when the next word consists of multiple subword tokens, we combine the probabilities of all of them.

This teacher policy allows us to bootstrap attention model learning, and once we have a reasonably good attention (and thus reward function), we can start training the activation policy.
We observed that this bootstrapping is critical for convergence, without it the model fails to learn meaningful attention and activation functions: 
	attention to the default model approaches 1 irrespective of the input while activation fluctuates randomly as it no longer affects the output.

\subsection{Component Model Training}
\label{section:component-training}

Component models are trained independently on their respective datasets.
We do not make assumptions about component model types, any combination of WFST and neural models is possible.
There only two requirements: that all models must share the vocabulary, 
	and that components must predict \verb|</s>| with high probability after seeing an unknown input sequence.
For neural component models, this can be achieved by adding random input sequences with \verb|</s>| labels, and
for WFST models, we add a \emph{dead state} which only generates \verb|</s>| and all unmatched transitions lead to this state.

\section{Experiments}
\label{section:experiments}
In this section, we evaluate performance of our model on two tasks: perplexity on an English Wikipedia dataset, 
	and word error rate (WER) reduction on an ASR 20-best rescoring task using personalized model trained on anonymized transcriptions of interactions with a voice assistant.

\subsection{Model Parameters}
In both experiments in this section, we use the same structure for our model:

\begin{itemize}
	\item The default model is an LSTM model comprising of two 300-unit layers, and a skip connection~\cite{deepres2016} over both LSTM layers. 
	The input is 300-dimension subword embedding ($\context$) learned with the model, and the output is a softmax layer.
	\item The context encoder is a single layer LSTM with 256 units layer and 0.2 dropout. The input embeddings are shared with the default model and are frozen during the compositional LM training.
	\item Component embeddings ($\compemb$) have dimensionality of 256.
	\item The activation model is a 2-layer LSTM with 128 units, the LSTM's output is projected into a scalar.
	The input to activation model has dimensionality of $514 = 256+256+1+1$ (see Eq.~\ref{eq:activation}). We use 0.2 dropout between LSTM layers.
	\item Finally, the attention model is a single 128-unit LSTM layer with its output projected into a scalar.
	The attention model's input has $514$ dimensions (see Eq.~\ref{eq:attention}). We add 0.2 dropout before and after the LSTM layer.
\end{itemize}

For component models, we use WFST-based representation which varies in size depending on the component.
The WFST is constructed as a union of entities, determinized and minimized under \verb|log semiring| \citep{mohri1997wfst}.
In Wikipedia experiments, each entity is weighted according to its frequency in the training data, and in n-best rescoring models, the distribution is uniform.

\subsection{Training Procedure}
\label{section:training-procedure}
We use Adam optimizer~\citep{kingma2014adam} with 0.001 initial learning rate and an exponential decay of 0.7 per 1k updates.
We define an epoch as 800 updates.
For efficiency, we use truncated backpropagation with a chunk size of 16 and a batch size of 160.
Note that due to truncated backpropagation, the reward outside the current chunk will not be added in cumulative reward computation $G_{t}$ (Eq.~\ref{eq:reinforce}).
This introduces some noise but at the same time it reduces the variance of update, especially in long sentences.
The impact (reward) of a component's activation is limited to the span it generates, therefore, as long as the chunk size is significantly larger than the typical component's output, chunking should not have a negative impact.

First, we do ``pre-training'' of all parts of the compositional model except activation by using teacher activation only ($\lambda = 1$ in Eq.~\ref{eq:interpolated-policy}, meaning that all activation samples are drawn from the teacher model) and LL loss (Eq.~\ref{eq:LL}) for 5 epochs.
Then we run the main training routine for 20 epochs, alternating between LL and RL losses every 1-3 batches randomly\footnote{We found that alternating the losses yields slightly better results than their sum.}.
At the same time $\lambda$ (Eq.~\ref{eq:interpolated-policy}) is exponentially decayed at the rate of 0.8 per 1k updates.
Once $\lambda$ reaches 0.05, it is set to 0, meaning that all activation samples will come from the activation model.

\subsection{Wikipedia}
\label{section:wikipedia}
\vspace{-1mm}
Evaluating the impact of adding components poses a challenge: in order to be useful, the components have to contain information that the default model has not been exposed to.
We are not aware of any publicly available datasets that satisfy this property, therefore we simulate this by using an automatically tagged Wikipedia corpus with fine-grained entity tags \citep{ghaddar2018wiki}.
We split the corpus into train/dev/test partitions by document.
Component models are then built from entity mentions corresponding to "\verb|/location/city|", "\verb|/person|", and "\verb|/organization/company|"
entity types using Figer scheme \citep{liu2014figer} in the entire \emph{train} partition.
The default model is built using a random subset of 100,000 training documents.
We also filter out stopword mentions (pronouns) as well as entities onger than 3 words (some entities are quite long and contain entire subordinate clauses).
The WFST models have 0.7M, 1.2M, and 0.8M arcs, respectively.

The entire Wikipedia corpus contains almost 7M unique words, we segment them into subwords using 30k subword vocabulary.
Subwords ending with an underscore indicate end-of-word, and activations are only generated at word boundaries.
Figure~\ref{fig:example-sentence} shows a sentence from this corpus.

\begin{figure}[htbp]
\begin{center}
\includegraphics[width=1.0\linewidth]{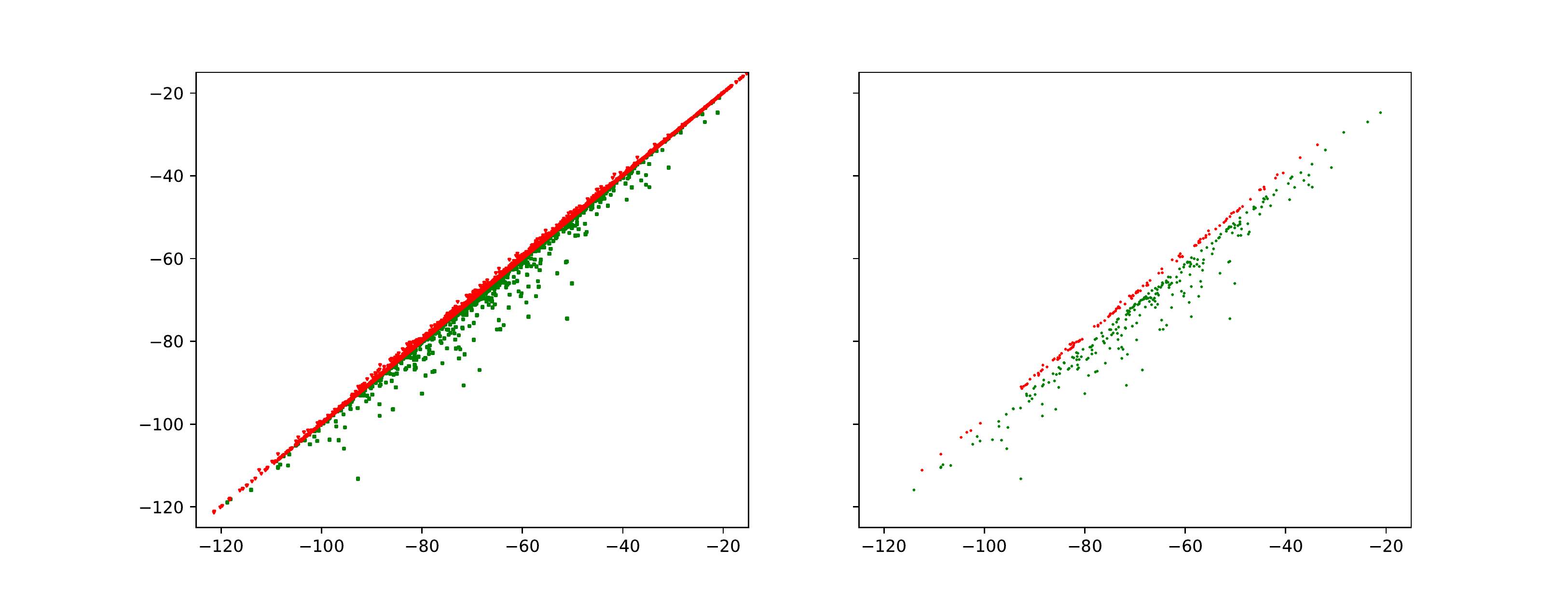}
\caption{Log-likehood (natural log) scatter plot of sentences: compositional model (X axis) vs. the default model (Y axis).
Green color indicates sentences with higher likelihood under the compositional model compared to the default, and red color signifies the opposite.
The first on the left contains all sentences up to 20 tokens.
On the right, we remove sentences whose scores differ by less that 1 to remove the clutter.
}
\label{fig:wiki-log-likelihood}
\end{center}
\end{figure}

We compute perplexity on a subset of 1,000 documents from test partition.
This test set contains 30k sentences with 970k subwords.
The perplexity of the default model is 64.9 and the compositional model's perplexity is 64.0 (1.4\% reduction).
However, we expect our compositional model to make a significant impact over the default model only on a fraction of sentences 
	because most sentences do not contain aforementioned entity types.
	Besides, the default model should already model frequent entities well enough.
In Figure~\ref{fig:wiki-log-likelihood}, we plot log-likelihood of the default model vs. the composite model.
We show sentences up to 20 tokens to limit the range.

\subsection{N-best rescoring}
\label{section:rescoring}
In this section, we evaluate the impact of our model on scoring n-best ASR hypotheses.
The test set consists of anonymized transcriptions of interactions with a voice assistant, 
	and each utterances is associated with an anonymized user id, 
	and some user ids have personalized models of "\verb|contact names|" associated with them, used to improve recognition accuracy for "\verb|communication|" domain.
We use separate partitions for train/dev/test with disjoined sets of users.
Only a small fraction of all interactions belong to ``\verb|communications|'' domain, and even smaller fraction still involve contact names.
Therefore, we present results on ``All domains'' test set which represents general interactions, 
	and separately, on ``Communications'' domain test set.
Our goal is to improve the performance on interactions that do invoke contact names, while not regressing on the rest of the data.

In Table~\ref{tbl:ppl}, ``default component'' is trained on a large sample of interactions representative of "all domains".
The "compositional" model is trained on additional 0.8\% of data, half of which belongs to the "\verb|communication|" domain.\footnote{We did evaluate the default component trained on the additional data, but found no significant difference.}

The compositional model is trained with a single component (not counting the default) representing "\verb|contact names|" entity.
To simplify the training procedure, we used a ``unified entity'' model built from aggregated contact lists across all users in the training data. However, for evaluation we also report results with personalized entity.
We use a 10k subword token vocabulary for all models.

\begin{table}[htp]
\caption{Relative difference in perplexity compared to the default model.
Reductions (negative numbers) indicate improvements.}
\begin{center}
\begin{tabular}{|l|r|c|c|c|c|}
\hline
 & \#Utterances & \parbox[c]{1.5cm}{\begin{center}Default\\component\end{center}} & \parbox[c]{2.2cm}{\begin{center}Compositional\\(unified entity)\end{center}} & \parbox[c]{2.3cm}{\begin{center}Compositional\\(personal entity)\end{center}} \\
\hline
All domains							& 138,094 & - & -0.3\%  & 0.1\% \\
Communications						&   14,943 & - & -8.9\%  & -22.2\% \\
~ only w/ contact names 					&     6,574 & - & -20.6\% & -43.6\% \\
~ only w/ personal entities 				&     4,181 & - & -19.3\% & -51.8\% \\
\hline
\end{tabular}
\end{center}
\label{tbl:ppl}
\end{table}

In Table~\ref{tbl:ppl}, we compare perplexity of our models and the default model.
On the entire test set, the changes in perplexity are insignificant which indicates that our model does not cause regression on utterances where its component is not used.
On subsets of utterances where adding contact names is expected to make a difference, we do observe substantial reductions in perplexity.
Note that despite learning the component embedding using unified "\verb|contact names|", we observe better performance by using personalized entities.
This indicates that the component embedding learns a representation of \emph{entity type} rather than its content.

We also evaluate these models on an n-best rescoring task~\cite{rescorenlm2019}.
We rescore top 20 hypotheses generated by a hybrid CTC-HMM ASR model \citep{graves2006ctc}, trained on a large amount of anonymized transcriptions, using the default component as the baseline and compare that to rescoring with our proposed compositional model.
The results are presented in Table~\ref{tbl:wer}.
Improvements of compositional models over the default component are significant with at least $p<0.0001$.

\begin{table}[htp]
\caption{Relative difference in WER compared to 1-best of CTC-HMM model.
Reductions (negative numbers) indicate improvements.}
\begin{center}
\begin{tabular}{|l|c|c|c|c|c|}
\hline
 & \parbox[c]{1.5cm}{\begin{center}Default\\component\end{center}}& \parbox[c]{2.2cm}{\begin{center}Compositional\\(unified entity)\end{center}} & \parbox[c]{2.3cm}{\begin{center}Compositional\\(personal entity)\end{center}} & Oracle \\
\hline
All domains							& -3.0\% & -2.9\% & -3.0\% & -30.3\% \\
Communications	 					& -1.3\% & -2.5\% & -5.8\% & -38.8\% \\
~ only w/ contact names 					& -0.9\% & -2.3\% & -6.8\% & -40.5\% \\
~ only w/ personal entities 				& -0.8\% & -2.8\% & -10.0\% & -45.8\% \\
\hline
\end{tabular}
\end{center}
\label{tbl:wer}
\end{table}
\vspace{-2mm}
\section{Conclusions and Future Work}
\label{section:conclusions}
\vspace{-3mm}
In this paper, we proposed a novel method how to compose separately trained models, including personalized models, with a general generative language model.
We showed that our method is effective at learning the composition directly from data without relying on annotations. 
While we evaluate our approach on language modeling tasks, we believe our approach can be applied to many sequence-generating applications in natural language processing.
In the future, we plan to integrate our model directly into ASR decoder using end-to-end models such as LAS \citep{chan2015las} and RNN-T \citep{graves2012rnnt}.
We also want to explore other applications, such as machine translation.

%

\bibliography{compositional_nlm}
\bibliographystyle{acl_natbib}

\end{document}